%% file: iclr2027_conference.tex
\documentclass{article} %
\usepackage{iclr2027_conference,times}

\input{math_commands.tex}

\usepackage{hyperref}
\usepackage{url}

\usepackage{graphicx}
\usepackage{wrapfig}  %
\usepackage{subcaption}
\usepackage{multirow}
\usepackage{adjustbox}
\usepackage{svg}
\usepackage{booktabs}
\usepackage[tableposition=bottom]{caption}
\usepackage{paralist}
\usepackage{colortbl}
\usepackage{amsmath}
\usepackage{enumitem}
\usepackage{xcolor}

\usepackage{pifont}
\newcommand{\cmark}{\ding{51}}% ✓
\newcommand{\xmark}{\ding{55}}% ✗

\newcommand{\boldparagraph}[1]{\par\vspace{0.0em}\noindent{\bf #1.}}

\title{RoGe: Novel View Synthesis via End-to-End \\ Implicit Reconstruction and Generation}

\author{Xiaolei Lang$^{1}$ \quad Ze Kang$^{2}$\thanks{Work done during an internship at Xiaomi EV.} \quad Zehao Huang$^{1}$ \quad Naiyan Wang$^{1}$ \\[4pt] \textsuperscript{1}Xiaomi EV \quad \textsuperscript{2}Northeastern University
\\[4pt]
\url{https://jerry-locker.github.io/roge/}
}

\iclrfinalcopy %
\begin{document}

\maketitle

\input{sections/0_abstract}
\input{sections/1_introduction}

\input{sections/2_related}
\input{sections/3_method}

\input{sections/4_experiments}
\input{sections/5_conclusion}

\bibliography{iclr2027_conference}
\bibliographystyle{iclr2027_conference}

\appendix
\newpage
\input{sections/x_appendix}

\end{document}

%% file: math_commands.tex
\usepackage{amsmath,amsfonts,bm}

\def\eqref#1{equation~\ref{#1}}
\def\1{\bm{1}}

\DeclareMathAlphabet{\mathsfit}{\encodingdefault}{\sfdefault}{m}{sl}
\SetMathAlphabet{\mathsfit}{bold}{\encodingdefault}{\sfdefault}{bx}{n}

%% file: sections/0_abstract.tex
\begin{center}
    \centering
    \captionsetup{type=figure}
    \includegraphics[width=0.85\textwidth]{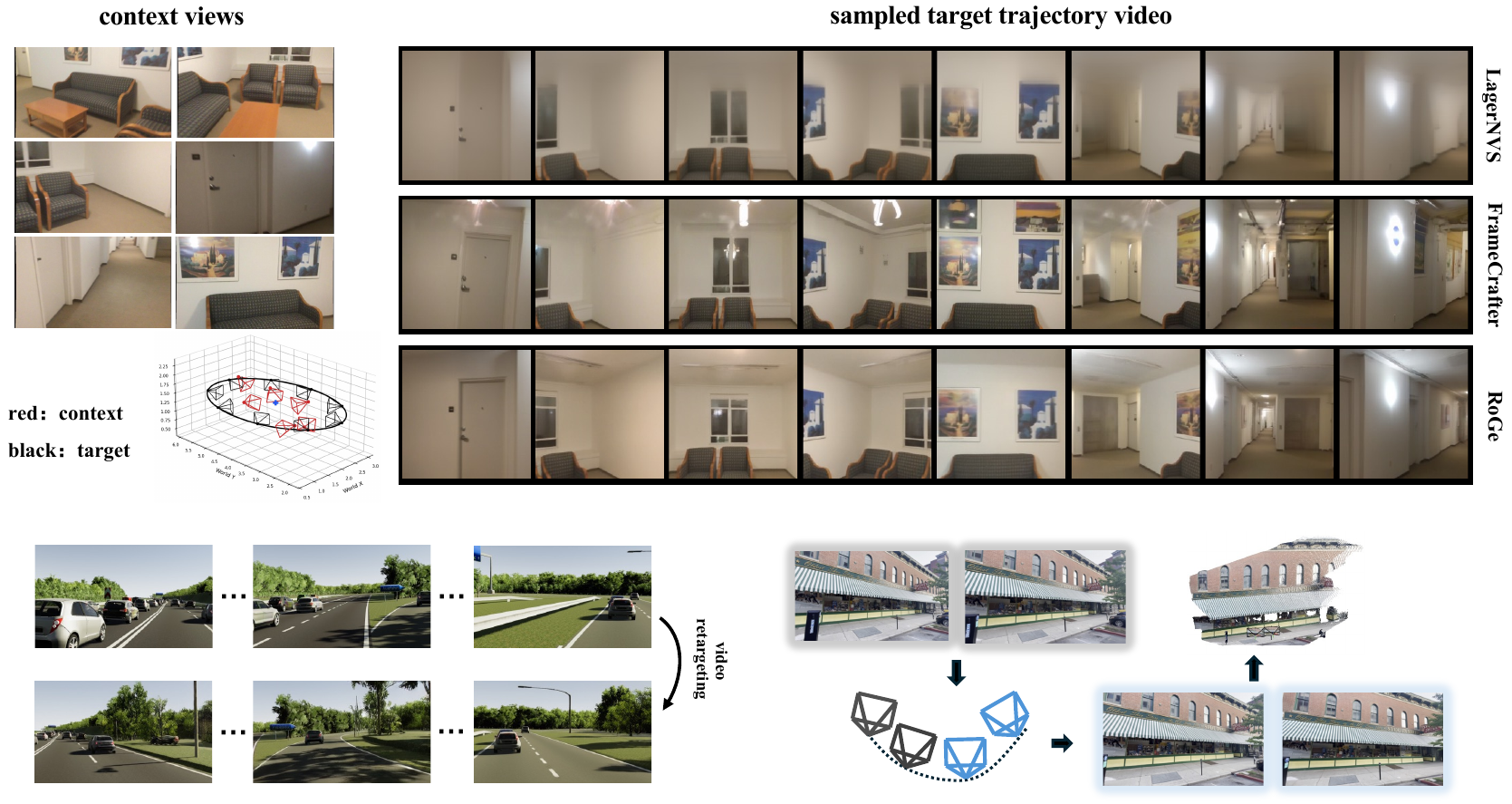}
    \captionof{figure}{\textbf{RoGe}, a unified reconstruction and generation framework for sparse novel view synthesis. Given sparse views and arbitrary camera trajectories, it lifts per-view implicit geometry into a video diffusion model to synthesize temporally and geometrically coherent videos with high visual fidelity, without any explicit 3D representation. (1) Top: roaming within the scene anchored by sparse views. (2) Bottom left: extending RoGe to video retargeting in the 4D world. (3) Bottom right: reconstructing geometrically accurate point clouds from the synthesized novel views.} 
    \label{fig:teaser}
\end{center}
\vspace{-1mm}

\vspace{-1mm}
\begin{abstract}
\vspace{-1mm}
Novel view synthesis from sparse inputs requires both geometric grounding from the observed views and generative priors of unobserved regions, motivating recent hybrid methods that combine reconstruction and generation. However, existing methods bridge the two with rendered images or explicit 3D representations such as point maps or 3D Gaussians. Generation is thus conditioned on a lossy and imperfect projection of the scene, inheriting its errors, and reconstruction receives no signal from generation to correct them. We present RoGe, an end-to-end unified reconstruction and generation framework that removes this explicit bridge. It targets roaming within a scene anchored by sparse views: given a few posed images and a camera trajectory, it synthesizes a temporally coherent video along that trajectory. From the sparse input views, RoGe builds an implicit scene representation with a feed-forward reconstruction model, and queries it with camera rays to obtain per-view geometric features. These features are injected into a video diffusion model as conditioning, without any explicit 3D intermediate. Both modules are trained jointly, so the generation objective directly shapes its own geometric conditioning. We conduct extensive experiments, where RoGe outperforms reconstruction-based, generation-based, and hybrid baselines in terms of image-level quality and video-level temporal and geometric consistency. Ablations confirm that ray-queried implicit features outperform both raw reconstruction tokens and rendered RGB as conditioning, and that joint training brings further gains.
\end{abstract}
\vspace{-2mm}

%% file: sections/1_introduction.tex
\vspace{-3mm}
\section{Introduction}
\vspace{-3mm}

Novel view synthesis (NVS) enables free-viewpoint exploration of captured scenes. While dense captures can support high-fidelity reconstruction, many practical settings provide only a few calibrated images. In this paper, we study free-viewpoint exploration in this sparse regime, which we term \textbf{\textit{roaming within a scene anchored by sparse views}}: given a sparse set of posed observations and a user-specified camera trajectory, the goal is to generate an ordered sequence of novel views that is visually realistic, temporally coherent, and consistent with the observed scene. This setting is inherently ill-posed, since large portions of the scene may be occluded or entirely unobserved. A successful method must therefore preserve the geometric structure and appearance supported by the input views, follow the requested camera motion, and plausibly complete missing content without introducing cross-view drift.

Existing methods for NVS largely fall into two paradigms. Reconstruction-based methods, including per-scene optimization~\cite{mildenhall2021nerf, kerbl20233d} and feed-forward models~\cite{jiang2025anysplat, ye2026yonosplat}, offer precise camera control and strong multi-view consistency, but struggle with unseen regions and extrapolation of poses, often resulting in blurring, floaters, and holes. Generation-based methods~\cite{gao2024cat3d,zhou2025stable,wu2026novel}, in contrast, leverage diffusion priors to synthesize unobserved areas under large viewpoint changes, yet lag behind in geometric consistency and camera controllability.
 
The complementary strengths have inspired efforts to bridge reconstruction and generation. One line of work~\cite{wu2024reconfusion, yu2024viewcrafter, wu2025difix3d+} augment per-scene optimization with diffusion-based inpainting, using generated views to extend coverage beyond observed regions. While effective, such a strategy is often limited by inconsistencies between the generated and input views. Another line, \cite{wu2025difix3d+, ren2025gen3c, yang2026neoverse} first reconstructs the 3D scene, and then render or project it to condition a generative model. Reconstruction and generation are thus fully decoupled. Generation only sees a lossy and imperfect projection of the scene and inherits its errors, while reconstruction receives no signal from generation to correct them.

In this work, we present RoGe, an end-to-end unified reconstruction and generation framework for NVS. RoGe extracts per-view geometric features from the implicit scene representation built from the intermediate tokens of a feed-forward reconstruction model and injects them into a video diffusion model as conditioning, without any explicit 3D intermediate representations. Crucially, the two models are trained end-to-end, enabling direct information flow between them and allowing the generative model to guide the learning of geometric representations. Experiments across both image-level and video-level metrics demonstrate that RoGe achieves superior visual quality, geometric consistency, and camera controllability. Main contributions can be summarized as follows:

\begin{itemize}[leftmargin=*, nosep]
    \item We propose RoGe, a joint reconstruction and generation framework for NVS that connects a feed-forward reconstruction network with a video generation model in an end-to-end manner, so that the generation model can directly consume the geometric representation it is conditioned on.
    \item We condition the generation part on per-view geometric features obtained by querying the implicit scene representation with camera rays. Ablations show that such ray-queried features are more effective than raw reconstruction tokens and rendered or decoded RGB maps.
    \item We conduct comprehensive experiments, demonstrating that our method synthesizes novel trajectory videos with high visual quality, strong geometric consistency, and precise camera controllability, surpassing existing reconstruction-based, generation-based, and hybrid methods.
\end{itemize}

%% file: sections/2_related.tex
\vspace{-4mm}
\section{Related Work}
\vspace{-3mm}

\boldparagraph{Scene Reconstruction}
Classical Structure-from-Motion (SfM) recovers sparse geometry through per-scene optimization, which is computationally expensive and brittle under sparse inputs~\cite{schonberger2016structure}. Feed-forward methods enable efficient multi-view reconstruction in a single pass~\cite{wang2024dust3r, leroy2024grounding, wang2025vggt, wang2026vggt, wang2026pi, keetha2026mapanything, lin2025depth}. However, both of these approaches primarily output point clouds, which are insufficient for high-fidelity NVS. Neural Radiance Fields~(NeRF)~\cite{mildenhall2021nerf} and 3D Gaussian Splatting~(3DGS)~\cite{kerbl20233d} instead provide renderable scene representations. 3DGS in particular offers fast optimization and real-time rendering at competitive visual quality, and has become a dominant choice for reconstruction-based NVS. Recently, feed-forward 3DGS has shown increasingly promising results, often outperforming optimization-based 3DGS under sparse inputs~\cite{charatan2024pixelsplat, chen2024mvsplat, xu2025depthsplat, ziwen2025long, ye2025no, jiang2025anysplat, ye2026yonosplat}. Departing from explicit representations, LagerNVS~\cite{szymanowicz2026lagernvs} shows that the intermediate tokens of VGGT~\cite{wang2025vggt} already encode sufficient scene geometry to render novel views directly. Despite strong multi-view consistency and precise camera control, these reconstruction-based methods, including LagerNVS, lack a generative prior and thus produce blur and artifacts in regions unobserved by the sparse inputs.

\boldparagraph{Video Generation}
Video generation models have demonstrated impressive capabilities in generating open-domain videos~\cite{blattmann2023stable, zheng2024open, yang2025cogvideox, kong2024hunyuanvideo, wan2025wan, agarwal2025cosmos, lin2026towards}. On top of these models, camera-controllable approaches generate videos from a single image or text prompt along user-specified trajectories~\cite{wang2024motionctrl, he2024cameractrl, he2025cameractrl, zhang2026unified, zhao2026cameranoise}. Another line of work focuses re-rendering existing videos along new trajectories~\cite{yu2025trajectorycrafter, jeong2025reangle, bai2025recammaster, bai2025syncammaster}. More closely related to classical NVS, CAT3D~\cite{gao2024cat3d}, SEVA~\cite{zhou2025stable} and FrameCrafter~\cite{wu2026novel} leverage multi-view or video diffusion models to synthesize novel views from sparse observations. While generation-based NVS methods excel at extrapolating unobserved content from sparse inputs, they generally lag behind reconstruction-based approaches in geometric consistency and precise camera control.

\boldparagraph{Reconstruction and Generation}
Given the complementary nature of reconstruction and generation, an increasing number of hybrid approaches have emerged. One family aligns video diffusion with feed-forward 3D models in the feature space to generate geometrically consistent videos and 3D quantities~\cite{huang2025jog3r, wu2026geometry, dai2026fantasyworld, huang2026gen3r}. These methods target world modeling from a single image or text prompt, where the geometry is imagined rather than observed. Our task instead starts from sparse posed views, whose geometry can be recovered and used to ground video generation. Several works~\cite{wu2024reconfusion, yu2024viewcrafter, liu20243dgs, wu2025genfusion} retain per-scene optimization and use generated views as additional supervision, but inconsistencies between the generated and input views limit further gains. Two-stage methods~\cite{wu2025difix3d+, ren2025gen3c, yang2026neoverse, kang2026geonvs} instead recover an explicit 3D representation first and then use it as a condition in generative models, so that reconstruction and generation remain decoupled. Our method requires neither per-scene optimization nor an explicit 3D intermediate representation. Geometric features derived from a feed-forward reconstruction network are directly injected into the video generation model, and these two modules are trained end-to-end so that the generation objective shapes its own geometric conditioning.

%% file: sections/3_method.tex
\vspace{-3mm}
\section{Method}
\vspace{-3mm}

Given a sparse set of posed context observations $\mathcal{S}=\{(\mathbf{I}_i,\mathbf{K}_i,\mathbf{T}_i)\}_{i=1}^{M}$ of a static scene, our goal is to synthesize a video along a target camera trajectory $\mathcal{T}=\{(\mathbf{K}_j,\mathbf{T}_j)\}_{j=1}^{N}$. Here, $\mathbf{I}\in\mathbb{R}^{3\times H\times W}$ denotes an image, while $\mathbf{K}$ and $\mathbf{T} = [\mathbf{R}\mid\mathbf{t}] \in SE(3)$ denotes its camera intrinsics and camera-to-world pose, respectively, where $\mathbf{T}$ consists of a rotation $\mathbf{R}\in SO(3)$ and a translation $\mathbf{t}\in\mathbb{R}^3$. Without loss of generality, the first view in $\mathcal{S}$ is set as the world frame and the translations of both $\mathcal{S}$ and $\mathcal{T}$ are normalized by the scale factor derived from the maximum translation magnitude in $\mathcal{S}$.

\begin{figure}[t]
    \centering
    \includegraphics[width=0.92\linewidth, trim={0mm 0cm 0 0cm},clip]{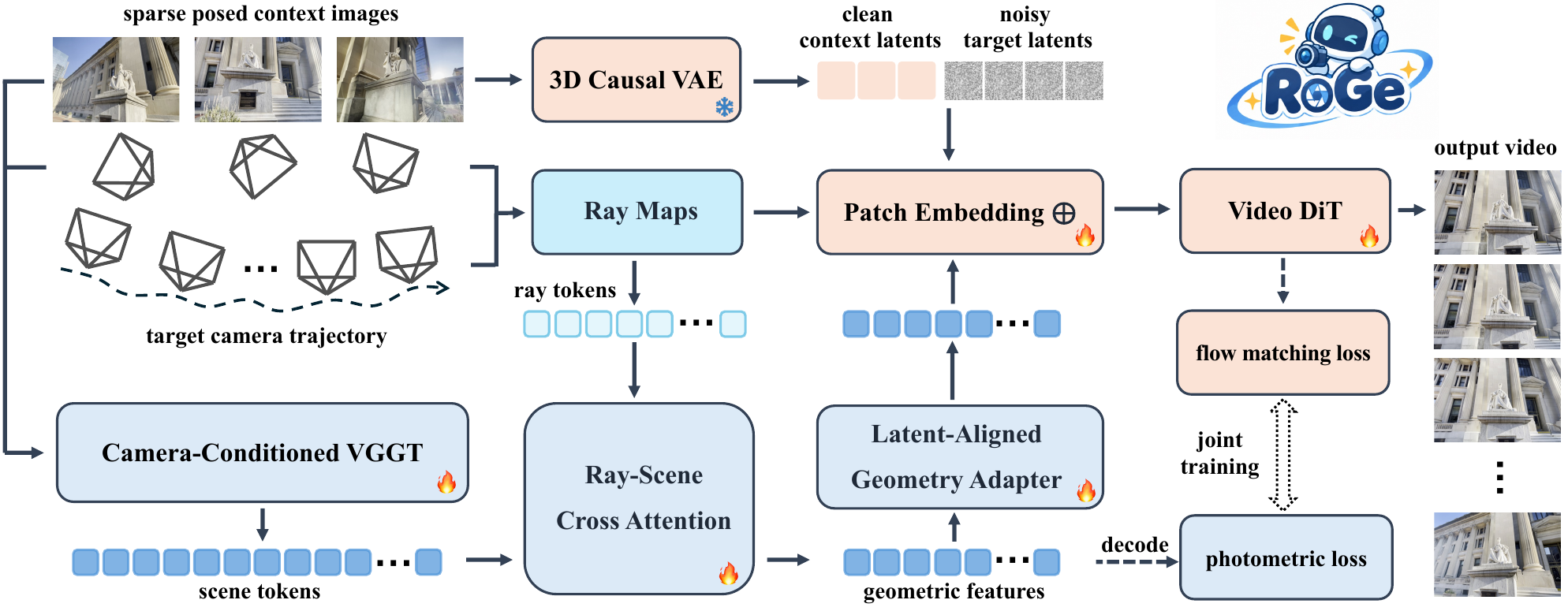}
    \caption{Overview of RoGe.  Given sparse posed images and a target camera trajectory, the camera-conditioned VGGT builds an implicit scene representation. Camera rays query this representation to extract geometric cues, which are injected into the video diffusion model to synthesize a photorealistic novel trajectory video. The reconstruction and generation models are trained end-to-end.} 
    \label{fig:pipeline}
    \vspace{-5mm}
\end{figure}

Sparse inputs leave large portions of the scene unobserved, making the problem inherently both geometric and generative. Neither reconstruction nor generation alone is sufficient to achieve optimal performance. We therefore bring these complementary paradigms together in a unified, differentiable framework shown in Fig.~\ref{fig:pipeline}. Specifically, we build an implicit scene representation from the sparse context views with a feed-forward 3D reconstruction model, and query it with per-view rays to output geometry-aware features for the target trajectory. We then inject these features into a pretrained video generation model as conditions, and optimize both modules jointly in an end-to-end manner. In the following, we first introduce two preliminaries, the Pl\"ucker ray representation and latent flow matching, and then elaborate the above pipeline in Sec.~\ref{sec:impl_recon}--\ref{sec:joint_train}.

\boldparagraph{Pl\"ucker Ray Representation}
We adopt Pl\"ucker ray map~\cite{plucker1865new, zhang2024cameras} as the common camera representation for both reconstruction and generation. For a pixel $\boldsymbol{\rho} = (u,v)$ from a view, let $\tilde{\boldsymbol{\rho}}=(u,v,1)^{\top}$ denote its homogeneous coordinate. Given the camera-to-world pose $\mathbf{T} = [\mathbf{R}\mid\mathbf{t}]$, the
corresponding Pl\"ucker coordinate of the pixel ray is defined as:
\begin{equation}
    \mathbf{p}
    =
    \bigl[
        \mathbf{d}, \,
        \mathbf{o}\times\mathbf{d}
    \bigr]\in\mathbb{R}^{6}, \quad\
    \mathbf{o} = \mathbf{t}, \quad\
    \mathbf{d}
    =
    \frac{
        \mathbf{R}\mathbf{K}^{-1}
        \tilde{\boldsymbol{\rho}}
    }{
        \left\|
        \mathbf{R}\mathbf{K}^{-1}
        \tilde{\boldsymbol{\rho}}
        \right\|_2
    },
\end{equation}

where $\mathbf{o}$ and $\mathbf{d}$ are the ray's origin and the normalized direction in the world frame. Stacking Pl\"ucker coordinates $\mathbf{p}$ over all pixels of the view yields a Pl\"ucker ray map $\mathbf{P}\in\mathbb{R}^{6\times H\times W}$. We retain two complementary encodings for such a ray representation: a tokenized ray map for querying the implicit scene and a lossless packed ray map directly supplied to the diffusion transformer (DiT)~\cite{peebles2023scalable}.

\boldparagraph{Latent Flow Matching}
We employ a pretrained video diffusion model operating in the latent space of a 3D causal Variational Autoencoder (VAE) as in~\cite{wan2025wan}. Let $\mathbf{X}_0$ denote the clean video latents and $\boldsymbol{\epsilon}\sim\mathcal{N}(0,\mathbf{I})$. 
% For a sampled point $\tau$ on the noise schedule, we construct:
Given a noise level $\sigma\in[0,1]$ sampled from the noise schedule, we construct the rectified flow path~\cite{liu2022flow, lipman2022flow}:
\begin{equation}\label{eq:rectified_flow}
    \mathbf{X}_{\sigma}
    =
    (1-\sigma)\mathbf{X}_0+\sigma\boldsymbol{\epsilon},
    \qquad
    \mathbf{V}^{*}
    =
    \boldsymbol{\epsilon}-\mathbf{X}_0,
\end{equation}
where the target velocity $\mathbf{V}^{*}$ is constant along the path. The diffusion transformer $\mathcal{D}_{\psi}$ predicts the velocity field that transports samples from Gaussian noise towards the data distribution:
\begin{equation}\label{eq:velocity}
    \hat{\mathbf{V}}_{\sigma}
    =
    \mathcal{D}_{\psi}
    \left(
        \mathbf{X}_{\sigma},\sigma;
        \mathcal{C}
    \right),
\end{equation}
where $\mathcal{C}$ denotes the conditioning signal detailed in Sec.~\ref{sec:recon_gen}. The model $\mathcal{D}_{\psi}$ is trained by regressing $\hat{\mathbf{V}}_{\sigma}$ to $\mathbf{V}^{*}$. During inference, the learned velocity field is integrated from $\sigma{=}1$ to $\sigma{=}0$, and the denoised latents are then decoded into output videos.

\vspace{-1mm}
\subsection{Implicit Reconstruction}\label{sec:impl_recon}
\vspace{-1mm}
Our implicit reconstruction module builds on LagerNVS~\cite{szymanowicz2026lagernvs}, a feed-forward NVS framework based on VGGT~\cite{wang2025vggt} without explicit 3D reconstruction.

\boldparagraph{Camera-Conditioned VGGT Reconstruction} 
VGGT is pretrained for 3D reconstruction tasks, estimating camera poses, depths, and point maps. Its intermediate tokens thus carry strong geometric priors and implicitly represent the 3D scene, from which we extract geometric cues for conditioning subsequent video generation. Specifically, we first use a two-layer Multi-Layer Perceptron (MLP) to project the intrinsics $\mathbf{K}$ and extrinsics $\mathbf{T}$ of each context frame in $\mathcal{S}$ into a 1024-dimensional token, which is added to the camera token that VGGT prepends to each view. The camera-conditioned VGGT network $\mathcal{R}_{\theta}$ then aggregates multi-view information through alternating local and global attention: $\{\mathbf{Z}_i\}_{i=1}^{M} = \mathcal{R}_{\theta}\left(\{\mathbf{I}_i, \mathbf{K}_i, \mathbf{T}_i\}_{i=1}^{M}\right)$. For each context view, we concatenate the tokens from the last local and global attention layers along the channel dimension, yielding 2048-dimensional scene tokens $\mathbf{Z}_i$. Unlike explicit methods, we do not further instantiate depth maps, point clouds, or 3D Gaussians. Instead, the tokens $\mathcal{Z} = \{\mathbf{Z}_i\}_{i=1}^{M}$ form an implicit, appearance-preserving scene representation that remains directly ‘renderable’ or queryable from arbitrary camera rays.

\boldparagraph{Ray Map Encoder}
We encode the Plücker ray map of each view in $\mathcal{S}$ and $\mathcal{T}$ as $\{\mathbf{Q}_i\}_{i=1}^{M+N} = \left\{\mathcal{E}_{\mathrm{ray}}(\mathbf{P}_i)\right\}_{i=1}^{M+N}$.
The 2D convolution $\mathcal{E}_{\mathrm{ray}}$ projects each $8\times8$ spatial patch into a 768-dimensional ray token, yielding a set of ray tokens $\mathbf{Q}_i \in \mathbb{R}^{768\times \frac{H}{8}\times \frac{W}{8}}$ for view $i$, with four learnable register tokens prepended to the ray tokens~\cite{darcet2024vision}. Such a dense representation preserves the spatial correspondence between each latent location and its associated camera ray.

\boldparagraph{Ray–Scene Cross-Attention}
Before interacting with the ray tokens, we linearly project the 2048-dimensional scene tokens $\mathcal{Z}$ to 768 dimensions. Then the ray tokens query the scene tokens through a stack of ray--scene transformer blocks. Each block first performs self-attention among the ray tokens of view $i$ and then exchanges information with the shared scene representation:
\begin{align}
    \tilde {\mathbf{Q}}_i^\ell
    &=
    {\mathbf{Q}}_i^\ell+
    \operatorname{SelfAttn}
    \bigl(
    {\mathbf{Q}}_i^\ell
    \bigr),\\
    {\mathbf{Q}}_i^{\ell+1}
    &=
    \tilde {\mathbf{Q}}_i^\ell
    +
    \operatorname{CrossAttn}
    \bigl(
        \tilde {\mathbf{Q}}_i^\ell,
        {\mathcal{Z}}^\ell,
        {\mathcal{Z}}^\ell
    \bigr),\\
    {\mathcal{Z}}^{\ell+1}
    &=
    {\mathcal{Z}}^\ell+
    \operatorname{CrossAttn}
    \bigl(
        {\mathcal{Z}}^\ell,
        \tilde {\mathbf{Q}}_i^\ell,
        \tilde {\mathbf{Q}}_i^\ell
    \bigr).
\end{align}
$\ell = 1, \ldots, L$. The first $L-1$ blocks use bidirectional cross-attention, allowing the scene tokens to be refined by the queried rays, while the final block performs ray-to-scene cross-attention only. After discarding the register tokens, the resulting ray tokens form a feature map ${\mathbf{\Phi}}_i \in \mathbb{R}^{768\times \frac{H}{8}\times \frac{W}{8}}$\footnote{In practice, the implicit reconstruction module operates on resized images. However, for simplicity of notation, we still use $H \times W$ to denote the image resolution in Sec.~\ref{sec:impl_recon}.}, which can be decoded to an RGB map to supervise the reconstruction module during joint training.

\vspace{-1mm}
\subsection{Reconstruction-Generation Fusion} \label{sec:recon_gen}
\vspace{-1mm}
After implicit reconstruction, we have the sparse context images $\{\mathbf{I}_i\}_{i=1}^{M}$, along with the Pl\"ucker ray maps $\{\mathbf{P}_i\}_{i=1}^{M+N}$ and the ray-queried geometric features $\{\mathbf{\Phi}_i\}_{i=1}^{M+N}$ for the context and target views. They respectively provide the appearance, the camera pose, and the implicit geometry that conditions the video DiT, forming $\mathcal{C}$ in Eq.~(\ref{eq:velocity}). We describe their construction below.

\boldparagraph{Hybrid VAE Representation}
Both the context images and the target video have to be encoded into the latent space of the VAE before entering the DiT. However, sparse context views form an unordered set, whereas the target views constitute a temporally coherent video. Encoding both with the same temporal strategy would either impose an artificial ordering on the context views or break temporal continuity in the output. We therefore adopt a hybrid latent representation: each context image $\mathbf{I}_i \in \mathcal{S}$ is independently encoded as a single-frame video following~\cite{wu2026novel}, while the target video is jointly encoded from the ground-truth frames $\{\mathbf{I}^{gt}_j\}_{j=1}^{N}$ available during training:
\begin{gather}
    {\mathbf{X}}^{\mathcal{S}}
    =
    [\mathcal{E}_{\mathrm{VAE}}(\mathbf{I}_1),\ldots,\mathcal{E}_{\mathrm{VAE}}(\mathbf{I}_M)]\in\mathbb{R}^{16\times M\times \frac{H}{8}\times \frac{W}{8}},\\
    {\mathbf{X}}^{\mathcal{T}}
    =
    \mathcal{E}_{\mathrm{VAE}}
    \left(
        [\mathbf{I}_1^{gt},\ldots,\mathbf{I}_N^{gt}]
    \right)
    \in\mathbb{R}^{16\times (1+\frac{N-1}{4})\times \frac{H}{8}\times \frac{W}{8}},\label{eq:tvae}
\end{gather}
where $[\cdot,\cdot]$ denotes concatenation along the temporal dimension. The clean training sequence is then formulated as ${\mathbf{X}}_0=[{\mathbf{X}}^{\mathcal{S}}, {\mathbf{X}}^{\mathcal{T}}]$. This hybrid representation preserves permutation invariance among the sparse context views while retaining the temporal structure of the causal video VAE for the target video. We additionally construct the appearance condition $\mathcal{Y} = \left[\mathbf{B}; \left[{\mathbf{X}}^{\mathcal{S}}, \mathbf{0}^{\mathcal{T}}\right] \right]$, where $[\cdot;\cdot]$ denotes concatenation along the channel dimension. $\mathbf{B}$ is a binary mask distinguishing context and target latent slots, and $\mathbf{0}^{\mathcal{T}}$ denotes zero-filled target latents.

\boldparagraph{Latent-Aligned Geometry Adapter}\label{sec:geom_adapter}
The implicit geometric features $\{\mathbf{\Phi}_i\}_{i=1}^{M+N}$ and VAE latents $\mathbf{X}_0$ differ in channel width, spatial resolution, and temporal organization. Hence we introduce a lightweight geometry adapter to bridge the two representations. For view $i$, $\mathbf{\Phi}_i$ is first projected from 768 to $C$ channels using a $1\times1$ convolution $\mathcal{E}_{\mathrm{geom}}$, and then bilinearly resampled to the latent resolution: ${\mathbf{\Phi}}^{\prime}_i = \operatorname{Resize}\left(\mathcal{E}_{\mathrm{geom}}(\mathbf{\Phi}_i)\right) \in\mathbb{R}^{C\times \frac{H}{8}\times \frac{W}{8}}$. In terms of temporal alignment, each of the $M$ context frames and the first target frame is replicated four times, and the remaining $N-1$ target frames are kept unchanged. Finally, every four consecutive frames are concatenated along the channel dimension, which yields the implicit geometry condition $\mathcal{G}\in\mathbb{R}^{4C\times(M+1+\frac{N-1}{4})\times \frac{H}{8}\times \frac{W}{8}}$.

\boldparagraph{Geometry-Grounded Diffusion Transformer}
We ground a pretrained video diffusion transformer~\cite{wan2025wan} $\mathcal{D}_{\psi}$ with both explicit camera rays and ray-queried geometric features. 

Similar to Sec.~\ref{sec:geom_adapter}, the Pl\"ucker ray maps of all views $\{\mathbf{P}_i\}_{i=1}^{M+N} \in \mathbb{R}^{6\times(M+N)\times H\times W}$ can be duplicated into $\mathbb{R}^{6\times(4M+4+N-1)\times H\times W}$, and then spatially rearranged using pixel unshuffle: $\mathcal{P} \in \mathbb{R}^{1536\times(M+1+\frac{N-1}{4})\times \frac{H}{8}\times \frac{W}{8}}$, where $6\times4\times8\times8 = 1536$. Pixel unshuffle preserves the complete per-pixel ray information while aligning with VAE latents. 

Together with $\mathcal{Y}$ and $\mathcal{G}$, $\mathcal{P}$ completes the conditioning signal $\mathcal{C}=\{\mathcal{Y},\mathcal{P},\mathcal{G}\}$ of the video diffusion transformer. We then patchify the noisy latents and the conditions, and fuse them together:
\begin{equation}
    \tilde{\mathbf{X}}_{\sigma}
    =
    \mathcal{E}_{\mathrm{patch}}^{1}
    \bigl(
        [\mathbf{X}_{\sigma}; \mathcal{Y}]
    \bigr)
    +
    \mathcal{E}_{\mathrm{patch}}^{2}
    (\mathcal{P})
    +
    \mathcal{E}_{\mathrm{patch}}^{3}
    (\mathcal{G}),
\end{equation}
where $\tilde{\mathbf{X}}_{\sigma} \in \mathbb{R}^{5120\times(M+1+\frac{N-1}{4})\times \frac{H}{16}\times \frac{W}{16}}$ and $5120$ is the hidden dimension of the DiT. ${\mathcal{E}}^1_{\mathrm{patch}}$, ${\mathcal{E}}^2_{\mathrm{patch}}$, and ${\mathcal{E}}^3_{\mathrm{patch}}$ denote 3D convolutional patch embeddings. The fused tokens $\tilde{\mathbf{X}}_{\sigma}$ are then processed by the subsequent DiT blocks. Note that temporal rotary embeddings are disabled on context latent slots for permutation equivariance, and are retained only along the target trajectory.

\vspace{-1mm}
\subsection{End-to-End Joint Training} \label{sec:joint_train}
\vspace{-1mm}

Our model is trained with reconstruction and generation supervision in a single
optimization step:
\begin{equation}
    \mathcal{L}
    =
    \mathcal{L}_{\mathrm{FM}}
    +
    \lambda
    \mathcal{L}_{\mathrm{photometric}}.
\end{equation}

\boldparagraph{Flow Matching Loss}
Given clean latents ${\mathbf{X}}_0=[{\mathbf{X}}^{\mathcal{S}}, {\mathbf{X}}^{\mathcal{T}}]$, we sample a noise level $\sigma$ and Gaussian noise $\boldsymbol{\epsilon}$ to construct noisy latents ${\mathbf{X}}_{\sigma}$ and the ground-truth velocity ${\mathbf{V}}^{*}$ as in Eq.~(\ref{eq:rectified_flow}). Since the context images are observed inputs, the loss is applied only to the target latent slots:
\begin{equation}
    \mathcal{L}_{\mathrm{FM}}
    =
    \mathbb{E}_{\mathbf{X}_0,\boldsymbol{\epsilon},\sigma}
    \left[
        w(\sigma) \cdot
        \left\|
            \mathcal{M}_{t}\odot
            \left(
                \hat{\mathbf{V}}_{\sigma}
                -
                {\mathbf{V}}^{*}
            \right)
        \right\|_2^2
    \right],
\end{equation}
where $\hat{\mathbf{V}}_{\sigma}=\mathcal{D}_{\psi}({\mathbf{X}}_{\sigma},\sigma;\mathcal{C})$ as in Eq.~(\ref{eq:velocity}), $\mathcal{M}_{t}$ selects the target latent slots with the operator $\odot$, and $w(\sigma)$ is the noise-level-dependent training weight.

\boldparagraph{Photometric Loss}
The reconstruction module projects the geometric features $\{\mathbf{\Phi}_i\}_{i=1}^{M+N}$ into RGB maps according to LagerNVS, supervised by a pixel-wise mean-squared error $\mathcal{L}_{\mathrm{photometric}}$ against the corresponding ground-truth images. This auxiliary loss works jointly with the generator to refine the reconstruction module while preventing it from forgetting its learned geometric priors. No depth, point map, or any other explicit 3D supervision is used in our training.

%% file: sections/4_experiments.tex
\vspace{-2mm}
\section{Experiments}
\vspace{-2mm}

\vspace{-1mm}
 \subsection{Experimental Setup}
\vspace{-1mm}

\boldparagraph{Implementation Details}
The camera-conditioned VGGT $\mathcal{R}_{\theta}$, ray map encoder $\mathcal{E}_{\mathrm{ray}}$, and $L=12$ ray–scene cross-attention blocks are initialized from LagerNVS~\cite{szymanowicz2026lagernvs}. We adopt Wan2.1-I2V-14B-480P~\cite{wan2025wan} as our video generation backbone, containing the VAE $\mathcal{E}_{\mathrm{VAE}}$, the patch embedding $\mathcal{E}_{\mathrm{patch}}^{1}$ for video latents and RGB conditions, and the DiT $\mathcal{D}_{\psi}$. The geometry adapter $\mathcal{E}_{\mathrm{geom}}$, which projects 768-dimensional ray-queried features to $C{=}384$ channels, and the patch embeddings $\mathcal{E}_{\mathrm{patch}}^{2}$ for ray maps and $\mathcal{E}_{\mathrm{patch}}^{3}$ for implicit geometry are randomly initialized. During training, we apply Low-Rank Adaptation (LoRA)~\cite{hu2021lora} with rank 32 to $\mathcal{D}_{\psi}$ and rank 8 to $\mathcal{R}_{\theta}$, $\mathcal{E}_{\mathrm{ray}}$, and the ray-scene cross-attention blocks. All parameters except the LoRA modules, $\mathcal{E}_{\mathrm{geom}}$, and the three patch embeddings are frozen. We randomly sample $M\in{3,\ldots,9}$ context views and use $N=81$ temporally ordered target frames. The same views are fed to the feed-forward reconstruction module at $288\times504$ and to the video generation module at $192\times336$ or $480\times832$. We first train the model at $192\times336$ for 15,000 steps on 192 H20 GPUs with a batch size of 1 and $\lambda_{\mathrm{render}}=0.1$. We then further fine-tune the model at $480\times832$ for 500 steps on 192 H20 GPUs, with the entire reconstruction module fully frozen.

\begin{figure}[t]
    % \vspace{-3mm}
    \centering
    \includegraphics[width=0.95\linewidth, trim={0mm 0cm 0 0cm},clip]{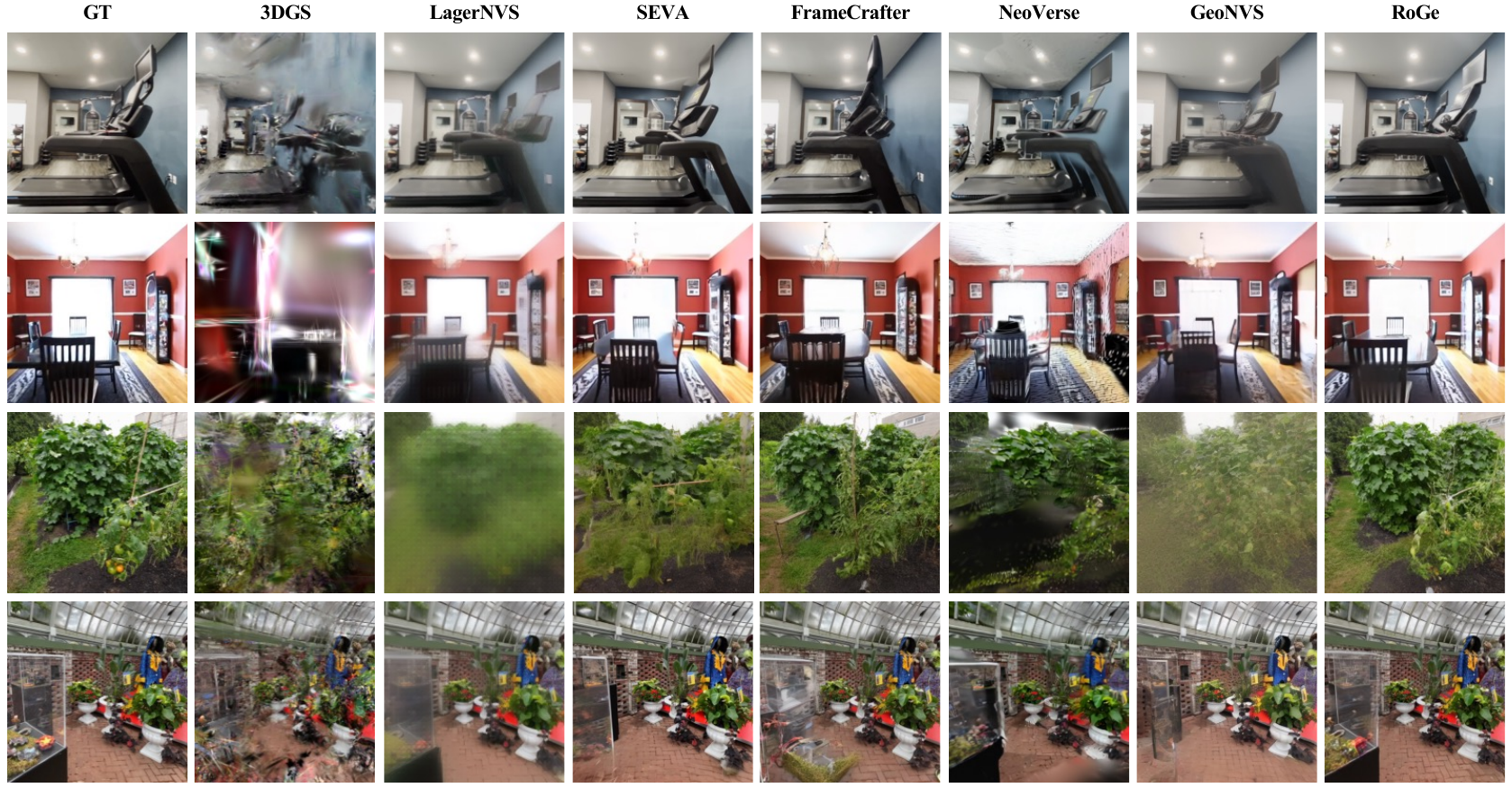}
    \caption{Qualitative NVS results across different methods with 3 or 6 sparse inputs.}
    \vspace{-6mm}
    \label{fig:re10k_dl3dv}
\end{figure}

\input{tables/1_re10k_dl3dv}

\boldparagraph{Datasets}
We train on RealEstate10K (RE10K)~\cite{zhou2018stereo} and  DL3DV~\cite{ling2024dl3dv}, which cover indoor and outdoor scenes with diverse camera trajectories and have been widely utilized for the training of feed-forward 3D models and camera-controllable video generation models. We follow CUT3R~\cite{wang2025continuous} for data preprocessing and official splits. After filtering, the training set contains 6,527 sequences from RE10K and 6,264 sequences from DL3DV. For quantitative evaluation, we use the 10 RE10K test sequences and 10 DL3DV test sequences selected by SEVA~\cite{zhou2025stable}. For cross-dataset generalization, we conduct experiments on ACID~\cite{liu2021infinite}, a large-scale aerial coastline dataset frequently used by prior works. We select 10 sequences with substantial viewpoint variation for evaluation, with an average of 474 frames per sequence, and follow SEVA to obtain the context and target splits. To cover the scene as comprehensively as possible with a minimal number of context views, we use 3, 6, and 9 context views for RE10K, DL3DV, and ACID, respectively. In addition, we evaluate NVS along extrapolated camera trajectories on ScanNet~\cite{dai2017scannet} to assess the capability of roaming.

\boldparagraph{Baselines}
We compare against representative state-of-the-art methods on NVS: 1) \emph{reconstruction-based}: 3DGS~\cite{kerbl20233d}, AnySplat~\cite{jiang2025anysplat}, YoNoSplat~\cite{ye2026yonosplat}, LagerNVS~\cite{szymanowicz2026lagernvs}; 2) \emph{generation-based}: SEVA~\cite{zhou2025stable}, FrameCrafter~\cite{wu2026novel}; 3) \emph{reconstruction-generation-combined}: Difix3D+~\cite{wu2025difix3d+} (facilitating 3DGS training or denoising rendered images), GEN3C~\cite{ren2025gen3c}, NeoVerse~\cite{yang2026neoverse}, GeoNVS~\cite{kang2026geonvs}. Since FrameCrafter is trained on only 1K DL3DV samples, we fine-tune it on the same data for the same number of steps as RoGe for a fair comparison. Note that NeoVerse is originally designed for synchronized video retargeting. We adapt it to sparse NVS by aggregating the predicted Gaussians from all frames for rendering.

\boldparagraph{Metrics}
Since we focus on trajectory NVS~\cite{zhou2025stable} rather than merely isolated novel views, we consider video-level metrics in addition to image-based ones. Image-level metrics include PSNR, SSIM, LPIPS, and DreamSim~\cite{fu2023dreamsim}. While for video-level evaluation, we adopt FID~\cite{heusel2017gans}, FVD~\cite{unterthiner2018towards}, TSED~\cite{yu2023long}, and MEt3R~\cite{asim2025met3r}. Specifically, FID and FVD assess the visual realism and temporal coherence of the generated video, while TSED and MEt3R focus on the geometric consistency. We further use VGGT-$\Omega$~\cite{wang2026vggt} to estimate the camera trajectory from the generated video and evaluate the RMSE of the translational and rotational Absolute Pose Error (APE), which simultaneously reflects visual quality, multi-view consistency, and camera controllability. We note that the ground-truth camera trajectories in RE10K, DL3DV, and ACID do not have an absolute scale. Therefore, for translational APE, we normalize the length of each ground-truth trajectory before computing the error. The ACID sequences contain nearly straight-line trajectories, for which the rotational component of $Sim(3)$ alignment can be ambiguous, making rotational APE unreliable. We therefore report rotational  Relative Pose Error (RPE) instead in ACID.

\input{tables/video_dl3dv}

\input{tables/nvs_acid}

\vspace{-1mm}
\subsection{Experimental Results and Analysis}
\vspace{-1mm}

\boldparagraph{Novel View Synthesis}\label{sec:nvs}
For each test sequence, image-based methods render, predict, or generate each non-context frame individually, whereas video-based methods generate non-context frames in 81-frame chunks, with each chunk produced in a single forward pass following their default inference settings. Since different methods produce images with different aspect ratios, we adopt a similar strategy to~\cite{wu2026novel} by cropping and resizing them to $192\times192$ to ensure identical FoV for final evaluation. As in Tab.~\ref{tab:1_re10k_dl3dv} and Fig.~\ref{fig:re10k_dl3dv}, per-scene optimization-based 3DGS struggles with sparse inputs, and Difix3D+ provides limited improvement by incorporating generative priors during optimization or directly refining the renderings. Feed-forward reconstruction methods such as YoNoSplat achieve stronger performance, but still suffer from holes and artifacts in regions not covered by the input views. Despite its high PSNR, LagerNVS tends to produce overly blurred results, especially under large viewpoint changes. On the other hand, generation-only methods such as SEVA and FrameCrafter exhibit strong capabilities of extrapolation and completion, but sometimes suffer from unstable camera control and geometric distortions. Hybrid methods, including GEN3C, NeoVerse, and GeoNVS, first reconstruct an explicit 3D representation and then obtain generation conditions from it through post-processing. The relatively complex post-processing pipeline plays a significant role in determining the final performance, while the two stages remain fully decoupled. As a result, they achieve only suboptimal performance. RoGe efficiently extracts geometric features from the implicit 3D representation to condition the video generation, while jointly training the reconstruction and generation modules end-to-end, achieving state-of-the-art visual quality across all the image-level metrics even on the generalization dataset (shown in Tab.~\ref{tab:3_acid}).

We also evaluate on the output 81-frame videos of each method. As reported in Tab.~\ref{tab:2_re10k_dl3dv} and Tab.~\ref{tab:3_acid}, FrameCrafter delivers good visual quality but struggles with geometric consistency. In contrast, reconstruction-based methods like LagerNVS display strong multi-view consistency and accurate camera control. By fully fusing reconstruction and generation, RoGe produces temporally coherent videos with high visual fidelity, consistent scene geometry, and accurate camera controllability.

\begin{table}[t]
    \centering
    \small
    \resizebox{0.35\linewidth}{!}{%
    \begin{tabular}{lccc}
        \toprule
        & MEt3R$\downarrow$ & APE-T$\downarrow$ & APE-R$\downarrow$ \\
        \midrule
        LagerNVS     & \textbf{0.059} & \underline{0.20} & \underline{3.52} \\
        SEVA         & 0.087 & 0.23 & 5.21 \\
        FrameCrafter & 0.153 & 0.57 & 7.96 \\
        GEN3C        & 0.114 & 1.48 & 10.61 \\
        NeoVerse     & \underline{0.069} & 0.25 & 4.92 \\
        GeoNVS       & 0.085 & 0.27 & 5.09 \\
        RoGe         & \textbf{0.059} & \textbf{0.15} & \textbf{3.12} \\
        \bottomrule
    \end{tabular}%
    }
    \caption{Quantitative comparison on the synthesized videos when roaming.}
    \label{tab:4_scannet}
\end{table}
\boldparagraph{Roaming in the Scene}
We further evaluate scene roaming along extrapolated trajectories that deviate from the trajectory along which the context views were captured. Tab.~\ref{tab:4_scannet} and Fig.~\ref{fig:teaser} present the results on ScanNet, showcasing RoGe's ability to extrapolate beyond the observed views while preserving the known scene geometry and  faithfully following the prescribed camera trajectory.

\boldparagraph{Extension for Video Retargeting}
We extend RoGe to video retargeting for 4D world modeling. Specifically, given a video with a known camera trajectory and a target trajectory for retargeting, we independently encode each posed context frame into scene tokens. We then apply ray–scene cross-attention to obtain ray-queried geometric features for each context frame and its corresponding target frame. The context video is instead encoded in the same way as the target video in Eq.~(\ref{eq:tvae}). We initialize the adapted model from the pretrained RoGe and further fine-tune it on VKitti~\cite{cabon2020virtual} for 5,000 steps using 16 H20 GPUs with a batch size of 1. Fig.~\ref{fig:teaser} shows an example on a test sequence, where a video captured from $15^\circ$ left is retargeted to $15^\circ$ right.

\vspace{-3mm}
\subsection{Ablation Studies}
\vspace{-3mm}

\begin{table}[t]
    \centering
    \small
    \resizebox{0.8\linewidth}{!}{%
    \begin{tabular}{clccccc}
        \toprule
        & Geometric Condition & Joint Train & Resolution & PSNR$\uparrow$ & SSIM$\uparrow$ & LPIPS$\downarrow$ \\
        \midrule
        \#1 & None (generation only)              & \xmark         & $192\times336$ & 18.49 & 0.517 & 0.156 \\
        \#2 & VGGT Patch tokens         & \xmark         & $192\times336$ & 17.98 & 0.477 & 0.168 \\
        \#3 & VGGT Patch tokens with camera-conditioned     & \xmark         & $192\times336$ & 18.83 & 0.541 & 0.147 \\
        \#4 & Rendered RGB                        & \xmark         & $192\times336$ & 19.36 & 0.564 & 0.141 \\
        \#5 & Predicted RGB                        & \xmark         & $192\times336$ & 19.49 & 0.573 & 0.139 \\
        \#6 & Ray-queried geometric features                & \xmark         & $192\times336$ & 20.07 & 0.608 & 0.133 \\
        \#7 & Ray-queried geometric features                & \cmark & $192\times336$ & 20.40 & 0.624 & 0.130 \\
        \#8 & Ray-queried geometric features                & \cmark & $480\times832$ & \textbf{21.50} & \textbf{0.704} & \textbf{0.104} \\
        \bottomrule
    \end{tabular}%
    }
    \caption{Ablation on geometric conditioning and joint training on DL3DV. \#1--\#6 freeze the reconstruction branch and differ only in what is injected into the video generation module. \#7 additionally trains both modules end-to-end. \#8 further fine-tunes the model at a higher resolution of $480\times832$.}
    \label{tab:ablation}
    \vspace{-4mm}
\end{table}

We ablate on DL3DV to answer two questions: what geometric condition to inject into the video generation branch, and whether to train the two modules jointly. Results are reported in Tab.~\ref{tab:ablation}.

\boldparagraph{Geometric Condition}
First of all, naively injecting raw VGGT patch tokens (\#2) is worse than no geometry at all (\#1). The world frame and scale underlying VGGT tokens are generally inconsistent with those of the input context and target poses. Although one could attempt to align them via $Sim(3)$, VGGT pose estimation or the alignment itself might fail, and even successful alignment cannot re-express the VGGT tokens in the new world frame and scale. As a result, the diffusion model receives geometry and camera control that disagree with each other. We instead condition VGGT with the input cameras (\#3), so that its features are built directly in the given frame. This resolves the mismatch and brings a modest gain over \#1. 

We query per-view geometric features (\#6) from the tokens output by the camera-conditioned VGGT and further decode them into RGB maps (\#5) following LagerNVS. We also experiment with connecting the camera-conditioned VGGT to the GS head of World-Mirror~\cite{liu2025worldmirror}, training the model on RE10K and DL3DV to predict Gaussians and render them into RGB maps (\#4). In comparison, using RGB maps from either implicit or explicit reconstruction as the conditioning signal provides less benefit than directly using the implicit geometric features.

\boldparagraph{Joint Training}
With ray-queried features as the condition, training the reconstruction branch together with the generation branch (\#7) further improves all metrics. The generation objective can now shape the geometric features it is conditioned on, which the frozen reconstruction branch in \#1--\#6 cannot benefit from. After fine-tuning at a higher resolution (\#8), the visual quality is improved.

%% file: tables/1_re10k_dl3dv.tex
\begin{table*}[t]
\centering
\caption{Quantitative NVS results on RE10K and DL3DV. \textit{R} and \textit{G} denote reconstruction-only methods and generation-only methods, respectively. \textit{R+G} means reconstruction-generation-combined methods, where \textit{1} and \textit{2} represent generation-aided reconstruction and decoupled reconstruction-then-generation. \textit{3} tightly couples two modules end-to-end without any explicit 3D. Difix3D and Difix apply generative priors during 3DGS training and after rendering, respectively.}
\label{tab:1_re10k_dl3dv}
\resizebox{0.85\textwidth}{!}{%
\begin{tabular}{lccccccccc}
\toprule
\multirow{2}{*}{Method} & \multirow{2}{*}{Type} & \multicolumn{4}{c}{RE10K \, 3v} & \multicolumn{4}{c}{DL3DV \, 6v} \\
\cmidrule(lr){3-6} \cmidrule(lr){7-10}
& & PSNR$\uparrow$ & SSIM$\uparrow$ & LPIPS$\downarrow$ & DreamSim$\downarrow$ & PSNR$\uparrow$ & SSIM$\uparrow$ & LPIPS$\downarrow$ & DreamSim$\downarrow$ \\
\midrule
3DGS & R & 14.76 & 0.449 & 0.469 & 0.380 & 13.75 & 0.308 & 0.426 & 0.438 \\
AnySplat & R & 16.77 & 0.572 & 0.190 & 0.069 & 13.21 & 0.279 & 0.315 & 0.176 \\
YoNoSplat & R & 18.83 & 0.641 & 0.142 & 0.063 & 17.04 & 0.459 & 0.277 & 0.184 \\
LagerNVS & R & \underline{32.09} & \underline{0.955} & 0.031 & 0.036 & \underline{20.90} & \underline{0.639} & 0.193 & 0.122 \\
SEVA & G & 29.26 & 0.929 & 0.036 & 0.025 & 18.63 & 0.603 & 0.158 & 0.080 \\
FrameCrafter & G & 29.03 & 0.925 & 0.037 & \underline{0.022} & 19.85 & 0.609 & \underline{0.121} & \underline{0.057} \\
\midrule
Difix3D & R+G, 1 & 14.85 & 0.452 & 0.425 & 0.325 & 13.89 & 0.313 & 0.388 & 0.375 \\
3DGS + Difix & R+G, 2 & 14.42 & 0.427 & 0.461 & 0.358 & 13.45 & 0.276 & 0.421 & 0.413 \\
LagerNVS + Difix & R+G, 2 & 30.98 & 0.895 & \underline{0.027} & 0.034 & 19.74 & 0.574 & 0.181 & 0.119 \\
GEN3C & R+G, 2 & 23.55 & 0.816 & 0.096 & 0.043 & 12.78 & 0.256 & 0.434 & 0.269 \\
NeoVerse & R+G, 2 & 16.20 & 0.538 & 0.264 & 0.102 & 16.45 & 0.432 & 0.237 & 0.166 \\
GeoNVS & R+G, 2 & 27.89 & 0.913 & 0.049 & 0.047 & 18.78 & 0.619 & 0.155 & 0.077 \\
RoGe & R+G, 3 & \textbf{33.35} & \textbf{0.965} & \textbf{0.021} & \textbf{0.020} & \textbf{21.50} & \textbf{0.704} & \textbf{0.104} & \textbf{0.052} \\
\bottomrule
\end{tabular}
}
\vspace{-6mm}
\end{table*}

%% file: tables/video_dl3dv.tex
\begin{table}[t]
    \centering
    \small
    \resizebox{\textwidth}{!}{%
    \begin{tabular}{lcccccccccccc}
        \toprule
        & \multicolumn{6}{c}{RE10K \, 3v} & \multicolumn{6}{c}{DL3DV \, 6v} \\
        \cmidrule(lr){2-7} \cmidrule(lr){8-13}
        & FID$\downarrow$ & FVD$\downarrow$ & TSED$\uparrow$ & MEt3R$\downarrow$ & APE-T$\downarrow$ & APE-R$\downarrow$
        & FID$\downarrow$ & FVD$\downarrow$ & TSED$\uparrow$ & MEt3R$\downarrow$ & APE-T$\downarrow$ & APE-R$\downarrow$ \\
        \midrule
        LagerNVS     & 21 & \underline{76} & \underline{0.994} & \textbf{0.014} & \textbf{0.007} & 12.51 & 33 & 206 & \underline{0.977} & \textbf{0.062} & \textbf{0.005} & \underline{2.42} \\
        SEVA         & 15 & 78 & 0.987 & 0.019 & 0.018 & 20.99 & 23 & \underline{85}  & 0.972 & 0.076 & 0.011 & 5.38 \\
        FrameCrafter & \underline{14} & 83 & 0.969 & 0.018 & 0.030 & 28.37 & \underline{16} & 89 & 0.720 & 0.080 & 0.018 & 6.18 \\
        GEN3C        & 31 & 154 & \textbf{0.997} & 0.020 & 0.027 & 23.68 & 61 & 281 & 0.854 & 0.089 & 0.057 & 35.24 \\
        NeoVerse     & 69 & 324 & \textbf{0.997} & 0.023 & \underline{0.009} & \underline{10.69} & 59 & 184 & 0.919 & 0.079 & \underline{0.008} & 2.91 \\
        GeoNVS       & 24 & 92 & 0.986 & 0.021 & 0.018 & 23.63 & 29 & 136  & 0.928     & 0.083 & 0.021   & 11.63 \\
        RoGe         & \textbf{11} & \textbf{45} & \textbf{0.997} & \underline{0.015} & \underline{0.009} & \textbf{9.45} & \textbf{15} & \textbf{50}  & \textbf{0.991} & \underline{0.071} & \textbf{0.005} & \textbf{2.31} \\
        \bottomrule
    \end{tabular}%
    }
    \caption{Quantitative comparison on RE10K and DL3DV with video-level metrics.}
    \vspace{-5mm}
    \label{tab:2_re10k_dl3dv}
\end{table}

%% file: tables/nvs_acid.tex
\begin{table}[t]
\centering
\caption{Quantitative comparison on ACID with both image-level and video-level metrics.}
\label{tab:3_acid}
\resizebox{0.86\linewidth}{!}{%
\begin{tabular}{lcccccccccc}
\toprule
 & \multicolumn{10}{c}{ACID \, 9v} \\
\cmidrule(lr){2-11}
 & PSNR$\uparrow$ & SSIM$\uparrow$ & LPIPS$\downarrow$ & DreamSim$\downarrow$ & FID$\downarrow$ & FVD$\downarrow$ & TSED$\uparrow$ & MEt3R$\downarrow$ & APE-T$\downarrow$ & RPE-R$\downarrow$ \\
\midrule
LagerNVS     & 24.11 & 0.708 & 0.139 & 0.088 & 29 & 133 & \underline{0.994} & \underline{0.028} & \underline{0.032} & \underline{0.21} \\
SEVA         & 22.97 & 0.696 & 0.094 & 0.048 & \underline{12} & \underline{78} & 0.969 & 0.048 & 0.043 & 0.37 \\
FrameCrafter & \underline{24.58} & \underline{0.734} & \underline{0.081} & \underline{0.040} & \underline{12} & 88 & 0.973 & 0.031 & 0.046 & 0.38 \\
GEN3C        & 16.39 & 0.427 & 0.288 & 0.153 & 43 & 313 & 0.989 & 0.031 & 0.057 & 0.22 \\
NeoVerse     & 15.99 & 0.401 & 0.322 & 0.209 & 89 & 399 & \textbf{0.995} & 0.032 & 0.036 & 0.27 \\
GeoNVS       & 22.07 & 0.656 & 0.134 & 0.082 & 24 & 285 & 0.934 & 0.057 & 0.058 & 0.70 \\
RoGe         & \textbf{26.30} & \textbf{0.786} & \textbf{0.064} & \textbf{0.032} & \textbf{10} & \textbf{36} & \textbf{0.995} & \textbf{0.027} & \textbf{0.028} & \textbf{0.20} \\
\bottomrule
\end{tabular}%
}
\vspace{-5mm}
\end{table}

%% file: sections/5_conclusion.tex
\vspace{-6mm}
\section{Conclusion}
\vspace{-4mm}
In this work, we present RoGe, a unified reconstruction and generation framework for novel view synthesis. RoGe extracts per-view implicit geometric features from a feed-forward reconstruction model and injects them into a video generation model, tightly coupling geometric reconstruction with generation without an intermediate explicit 3D representation. End-to-end joint training enables effective information exchange between the two modules, allowing RoGe to synthesize temporally and coherent videos along arbitrary camera trajectories with strong geometric consistency and visual fidelity, including in unseen regions. Extensive experiments on public dataset demonstrate that RoGe surpasses state-of-the-art reconstruction-based, generation-based, and hybrid methods.

%% file: sections/x_appendix.tex
\begin{figure}[t]
    % \vspace{-3mm}
    \centering
    \includegraphics[width=\linewidth, trim={0mm 0cm 0 0cm},clip]{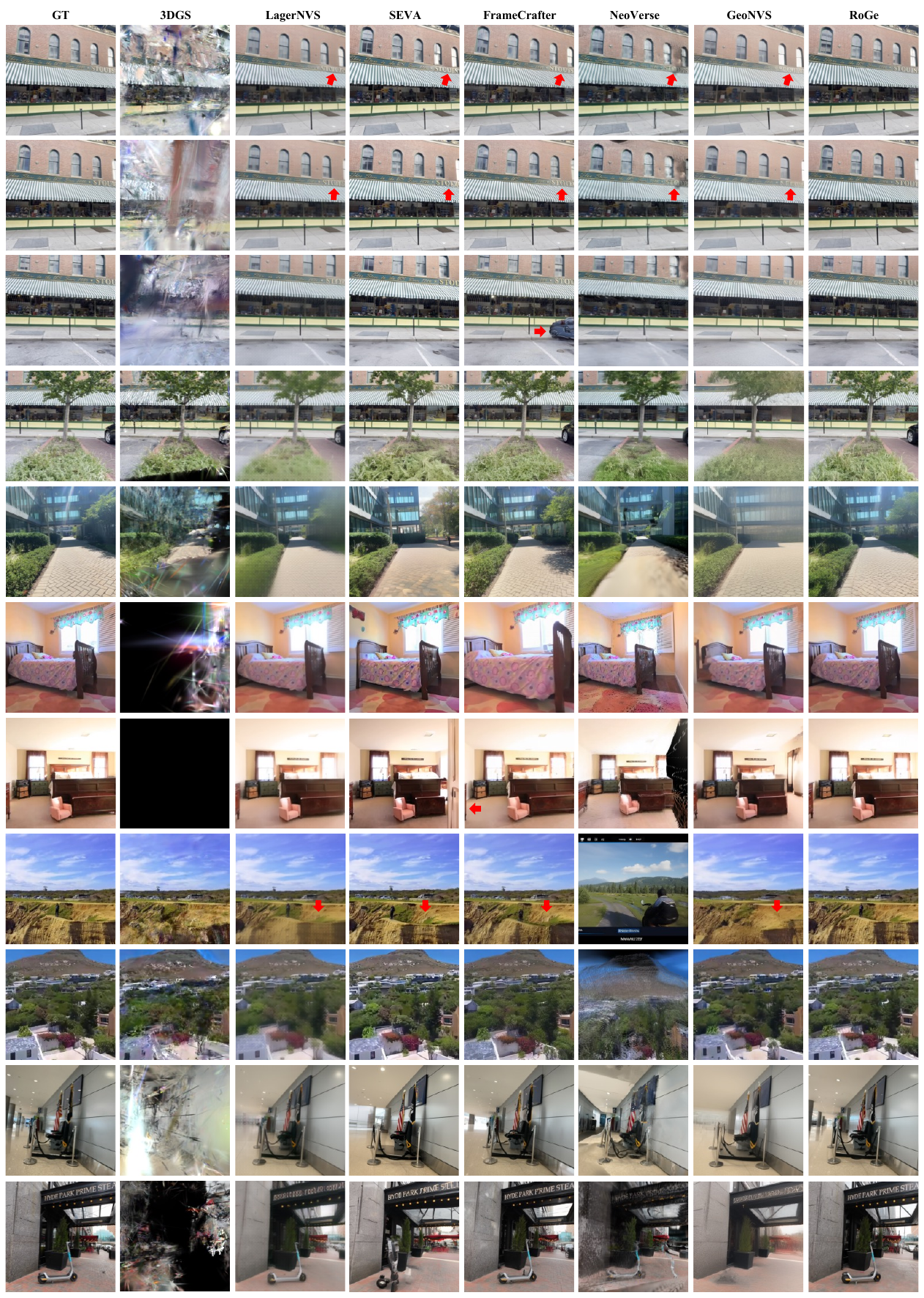}
    \caption{Qualitative results across reconstruction-based, generation-based, and hybrid methods.}
    % \vspace{-6mm}
    \label{fig:re10k_dl3dv}
\end{figure}

\begin{figure}[t]
    % \vspace{-3mm}
    \centering    \includegraphics[width=0.7\linewidth, trim={0mm 0cm 0 0cm},clip]{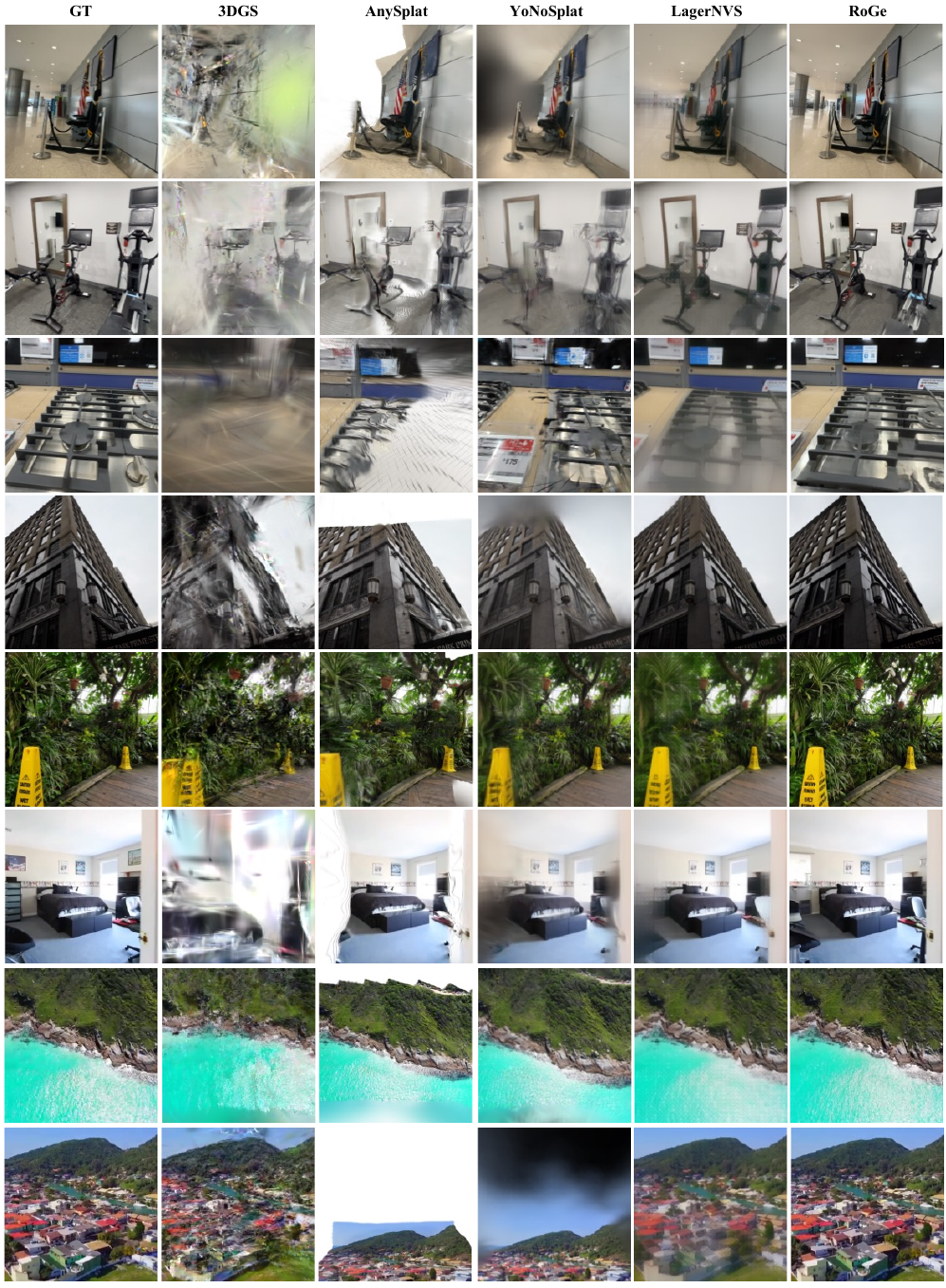}
    \caption{Qualitative NVS comparison with reconstruction-based methods.}
    % \vspace{-6mm}
    \label{fig:re10k_dl3dv}
\end{figure}

\begin{figure}[t]
    % \vspace{-3mm}
    \centering    \includegraphics[width=0.7\linewidth, trim={0mm 0cm 0 0cm},clip]{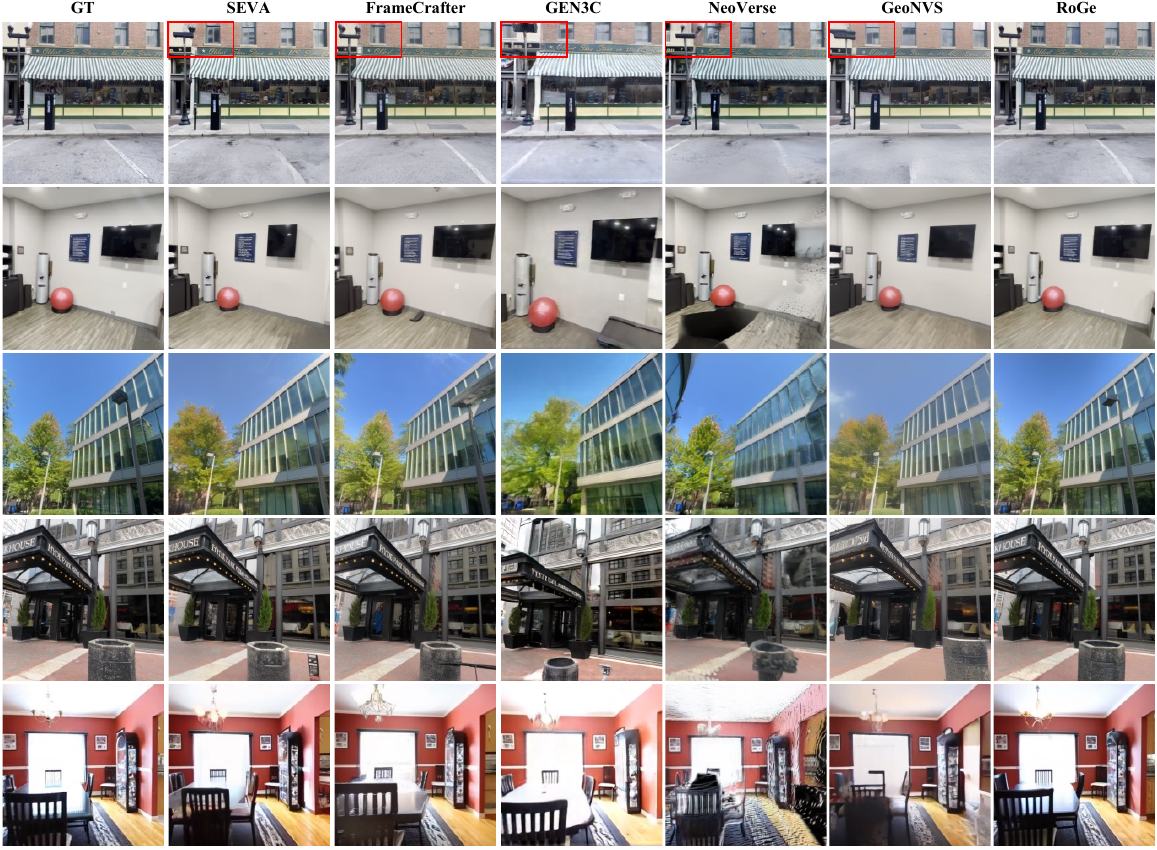}
    \caption{Qualitative NVS comparison with generation-based and hybrid methods.}
    % \vspace{-6mm}
    \label{fig:re10k_dl3dv}
\end{figure}